\documentclass[conference]{IEEEtran}
\usepackage{cite}
\usepackage{amsmath,amssymb,amsfonts}
\usepackage{algorithmic}
\usepackage{graphicx}
\usepackage{algorithm,algorithmic}
\usepackage{hyperref}
\hypersetup{hidelinks=true}
\usepackage{textcomp}
\usepackage{booktabs} 
\usepackage{bbding}
\usepackage{orcidlink}

\def\BibTeX{{\rm B\kern-.05em{\sc i\kern-.025em b}\kern-.08em
    T\kern-.1667em\lower.7ex\hbox{E}\kern-.125emX}}
\newcommand{\eat}[1]{}

\begin{document}
\title{Explainable Deep Radiogenomic Molecular Imaging for MGMT Methylation Prediction in Glioblastoma}

\author{
\IEEEauthorblockN{Hasan M Jamil\,\orcidlink{0000-0002-3124-3780} \Envelope }
		\IEEEauthorblockA{Department of Computer Science\\ University of Idaho, USA\\
			jamil@uidaho.edu}
}

\date{February 2025}

\maketitle

\begin{abstract}
Glioblastoma (GBM) is a highly aggressive primary brain tumor with limited therapeutic options and poor prognosis. The methylation status of the O6-methylguanine-DNA methyltransferase (MGMT) gene promoter is a critical molecular biomarker that influences patient response to temozolomide chemotherapy. Traditional methods for determining MGMT status rely on invasive biopsies and are limited by intratumoral heterogeneity and procedural risks. This study presents a radiogenomic molecular imaging analysis framework for the non-invasive prediction of MGMT promoter methylation using multi-parametric magnetic resonance imaging (mpMRI).

Our approach integrates radiomics, deep learning, and explainable artificial intelligence (XAI) to analyze MRI-derived imaging phenotypes and correlate them with molecular labels. Radiomic features are extracted from FLAIR, T1-weighted, T1-contrast-enhanced, and T2-weighted MRI sequences, while a 3D convolutional neural network learns deep representations from the same modalities. These complementary features are fused using both early fusion and attention-based strategies and classified to predict MGMT methylation status.

To enhance clinical interpretability, we apply XAI methods such as Grad-CAM and SHAP to visualize and explain model decisions. The proposed framework is trained on the RSNA-MICCAI Radiogenomic Classification dataset and externally validated on the BraTS 2021 dataset. This work advances the field of molecular imaging by demonstrating the potential of AI-driven radiogenomics for precision oncology, supporting non-invasive, accurate, and interpretable prediction of clinically actionable molecular biomarkers in GBM.
\end{abstract}


\section{Introduction}
\label{sec:introduction}
\IEEEPARstart{A}{ccording} to the National Brain Tumor Society, there are more than 100 different types of brain tumors and about one million Americans live with one \cite{ref1}. The mean survival rate is about 35.7\% of those with a malignancy \cite{ref2}. It is also widely known that brain cancers are one of the most complex cancers to treat because of the very sensitive surrounding structures as well as the relatively impermeable blood-brain barrier (BBB) to various chemotherapy drugs. Among the various types of brain cancers, glioblastoma multiforme is of particular interest here.

Glioblastomas (GBM) are the most common adult primary brain tumor and are aggressive, relatively resistant to therapy, and have a corresponding poor prognosis. They typically appear as heterogeneous masses centered in the white matter with irregular peripheral enhancement, central necrosis, and surrounding vasogenic edema, although molecular glioblastomas may appear indistinguishable from lower-grade astrocytoma, IDH-mutant \cite{ref3}. Glioblastoma (GBM) is an aggressive brain tumor where the O6-methylguanine-DNA
methyltransferase (MGMT) promoter methylation status is a crucial biomarker for prognosis and treatment response to temozolomide (TMZ) \cite{ref18} \cite{ref19}. Traditional assessment methods are invasive and prone to sampling errors due to intratumoral heterogeneity \cite{ref20}. Recent advancements in artificial intelligence, particularly radiomics and deep learning, offer promising non-invasive alternatives for predicting MGMT status from multi-parametric MRI (mpMRI) \cite{ref21}. Radiomics extracts quantitative features from images, while deep learning identifies complex patterns \cite{ref22},\cite{ref23}.The fusion of these approaches has shown enhanced accuracy, sensitivity, and specificity \cite{ref24}. However, the clinical translation of these methods is hindered by the 'black box' nature of complex algorithms and a lack of standardized datasets and robust external validation \cite{ref25}.

\section{Related Work}
Brain tumors, particularly glioblastoma multiforme (GBM), represent a significant global health challenge due to their complex and diverse nature, aggressive growth, and poor prognosis.Despite extensive research into their molecular and genetic characteristics, accurate diagnosis, prognosis, and treatment remain formidable tasks \cite{ref4}.

\subsection{Importance of MGMT Promoter Methylation and Radiogenomics}

O6-methylguanine-DNA methyltransferase (MGMT) promoter methylation is a crucial biomarker in glioblastoma, serving as both a prognostic indicator and a predictor of chemotherapy response, particularly to temozolomide (TMZ) \cite{ref5}.Patients with methylated MGMT disease generally progress later and survive longer, with a median survival rate of 22 months compared to 15 months for those with unmethylated MGMT \cite{ref6}. This methylation status affects the efficacy of chemotherapy by impacting the tumor cell's ability to repair DNA damage caused by alkylating drugs; if the promoter region is methylated, MGMT gene silencing occurs, increasing the tumor's vulnerability to chemotherapy \cite{ref7}.

Traditionally, determining MGMT promoter methylation status requires invasive surgical tissue sampling and subsequent molecular-genetic analysis (e.g., bisulfite modification and DNA sequencing, MSP, pyrosequencing). This process is labor-intensive, time-consuming (taking days to weeks), expensive, carries risks such as surgical morbidity and infection, and can be limited by sampling bias due to tumor heterogeneity. Consequently, there is an urgent need for non-invasive, accurate, and efficient methods to predict MGMT status \cite{ref5}. 
Radiogenomics has emerged as a promising frontier to address this need.It is defined as the integration of genomic and radiological data to correlate imaging information with biological processes like molecular pathways \cite{ref4}. Radiogenomics aims to achieve non-invasive precision medicine in cancer patients by allowing for virtual biopsies and augmenting predictive models with radiomic data to improve accuracy \cite{ref8}. It leverages advanced imaging techniques like Magnetic Resonance Imaging (MRI), Computed Tomography (CT), and Positron Emission Tomography (PET) to extract quantitative and qualitative imaging features that correlate with genetic and molecular tumor characteristics \cite{ref5}.

\subsection{Deep Learning and Fusion Approaches for MGMT Prediction}

Multiparametric MRI (mpMRI) is commonly used to identify and detect brain gliomas noninvasively, offering high-resolution 3D images without ionizing radiation \cite{ref9}.
Different MRI sequences provide complementary information:
\begin{itemize}
    \item FLAIR (Fluid-Attenuated Inversion Recovery) and T2-weighted (T2w) scans highlight the entire tumor and the tumor core regions, respectively \cite{ref10}.
    \item T1-weighted (T1w) and T1-weighted contrast-enhanced (T1wCE) scans emphasize enhancing tumor regions and cystic/necrotic components of the tumor core. These modalities (FLAIR, T1w, T1wCE, T2w) are crucial for the training of deep learning models for radiogenomic classification \cite{ref11}.
\end{itemize}
Recent advances in deep learning (DL) have driven notable progress in brain tumor research, particularly in radiogenomic analysis \cite{ref4}.DL models can learn hierarchical characteristics from input sequences, enabling automatic feature extraction and classification \cite{ref5}. Various DL architectures have been explored for MGMT prediction:
\begin{itemize}
    \item Convolutional Neural Networks (CNNs) are widely adopted for their success in image categorization and their ability to extract spatial features \cite{ref4}.
    \begin{itemize} %[label=$\circ$]
        \item Recurrent Neural Networks (RNNs) and Transformers focus on sequential and patch-specific information, respectively \cite{ref4}.
        \item Graph Neural Networks (GNNs) show potential for tasks involving complex and irregular objects like GBM tumors, by representing images as a graph structure and capturing both local and global features \cite{ref12}.
    \end{itemize}
\end{itemize}

Fusion-based approaches combine different data types or features to improve accuracy and mitigate data scarcity \cite{ref4}. Examples include:
\begin{itemize}
    \item Multi-fusion frameworks combining CNNs, RNNs, and transformers with contrastive learning to enhance feature representation and improve dark knowledge differentiation \cite{ref4}.
    \item Processing medical images through neural networks alongside a transformer backbone, merging outputs at various stages \cite{ref4}.
    \item Hybrid models that combine strengths of different machine learning models, such as K-nearest neighbor (KNN) and gradient boosting classifier (GBC), using soft voting criteria \cite{ref9}.
    \item Radiomics feature extraction combined with deep learning, often involving techniques like GLCM, HOG, and LBP \cite{ref13}.
    \item Multimodal information fusion from mpMRI sequences (T1w, T1wCE, T2w, FLAIR) to provide richer data representation and reduce ambiguity \cite{ref14}.
    \item Integrated pipelines that perform tumor segmentation as a first phase, then use the segmented tumor voxels as input for MGMT status classification \cite{ref15}.
\end{itemize}
Models like ResNet50 and DenseNet201 have achieved high accuracies (up to 100\% on test data) when utilizing transfer learning \cite{ref5}.Other studies report various performances, with AUC scores ranging from 0.577 to 0.90 depending on the model, dataset, and approach \cite{ref14}.

\subsection{Explainable AI (XAI)}

The "black box" nature of deep learning models poses a challenge for their adoption in clinical practice, as clinicians require transparency and trust in AI-driven decisions. Explainable AI (XAI) aims to bridge this gap by providing human-understandable insights into how AI models arrive at their predictions.

In the context of medical imaging, XAI techniques, such as Grad-CAM, LIME, and SHAP, are used to generate visual explanations (heatmaps) that highlight the regions in MRI scans most significant for a given classification \cite{ref16}. For MGMT methylation prediction, XAI can help clinicians understand which imaging phenotypes are driving the model's decision-making, facilitating human-machine collaboration \cite{ref17}.Some studies propose representing these explanations in natural language to further aid physicians' understanding \cite{ref16}.

\section{Methodology}
\label{sec:method}

\begin{figure*}
  \centering
  \includegraphics[width=\textwidth,keepaspectratio]{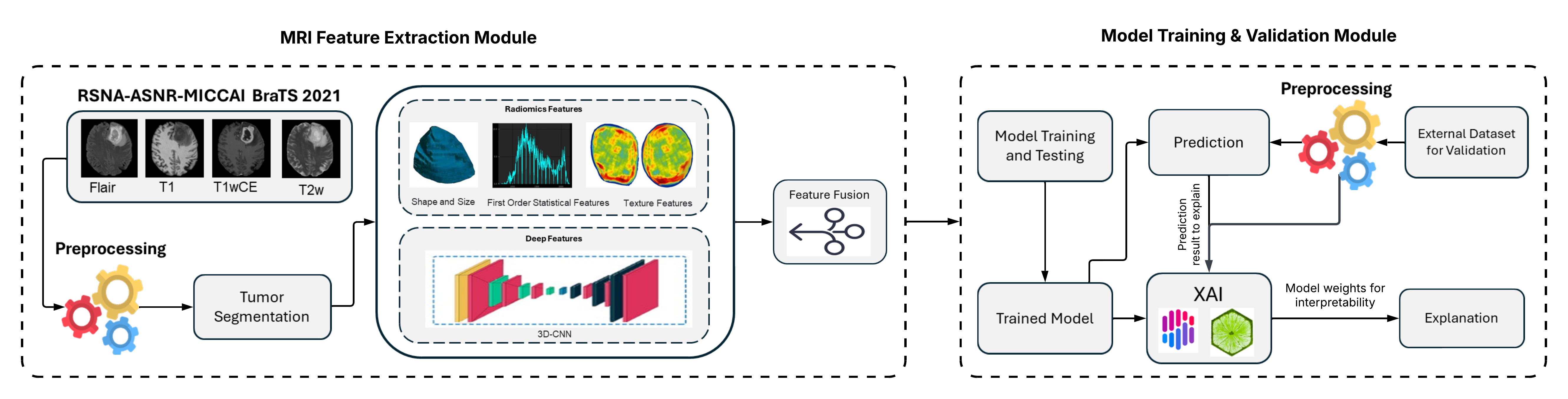}
  \caption{Overall pipeline. Left: MRI feature extraction (preprocessing, segmentation, radiomics, deep features, and feature fusion). Right: training, prediction, and interpretability.}
  \label{fig:pipeline}
\end{figure*}

As illustrated in Fig.~\ref{fig:pipeline}, our methodology is structured into two interconnected modules: an MRI Feature Extraction Module and a Model Training \& Validation Module. In the first stage, multi-parametric MRI scans from the BraTS 2021 dataset (FLAIR, T1, T1CE, and T2) undergo preprocessing and tumor segmentation to generate both binary masks and soft tumor probability maps. From these segmented regions, two complementary feature sets are derived: handcrafted radiomic descriptors (capturing shape, first-order statistics, and texture) and deep features learned through a 3D convolutional neural network. These features are fused to form a comprehensive representation of tumor biology. In the second stage, the fused representation is used to train a 3D ResNet-18 based classifier, validated on stratified 80/20 splits of BraTS 2021 and evaluated on an external cohort (UCSD-PTGBM) for cross-site robustness. Predictions are further interpreted using explainable AI (XAI) techniques, which leverage model weights and intermediate representations to generate clinically meaningful explanations. This design, implemented in our training code with ResUNet3D-DS for segmentation and a 3D ResNet-18 with radiomics fusion for classification, ensures that the pipeline not only achieves accurate MGMT methylation prediction but also supports interpretability and generalization across datasets.

\subsection{Datasets and Cohort Intersection}
\label{sec:datasets}

\subsubsection{RSNA-ASNR-MICCAI-BraTS-2021} 
The RSNA-ASNR-MICCAI-BraTS 2021 dataset \cite{dataset1}, a publicly available benchmark, was utilized for model training. This dataset comprises brain magnetic resonance imaging (MRI) scans of adult patients with glioma. Each patient's record includes four co-registered, three-dimensional (3D) MRI modalities: T2-weighted (T2), T1-weighted (T1), T1-weighted with contrast enhancement (T1CE), and Fluid Attenuated Inversion Recovery (FLAIR). The dataset also provides comprehensive metadata, including patient demographics and molecular characteristics such as MGMT promoter methylation status. Our study addresses two distinct but related tasks: 
\begin{enumerate}
\item \textbf{Task 1}: Tumor Segmentation. This task involves predicting pixel-wise segmentation masks of the glioma and its sub-regions. The segmentation masks are labeled according to the established BraTS classes: background, necrotic and non-enhancing tumor core (NET/NCR), peritumoral edema (ED), and enhancing tumor (ET). 
\item \textbf{Task 2}: MGMT Promoter Methylation Prediction. This task is formulated as a binary classification problem to determine the MGMT promoter methylation status of the tumor.
\end{enumerate} 
Using the official TCIA metadata and our custom data curation pipeline, we established three primary cohorts based on the availability of labels: a segmentation cohort for Task 1, an MGMT classification cohort for Task 2, and an {\em intersection cohort} comprising subjects with both valid segmentation masks and MGMT labels. The model for Task 1 was trained on the full segmentation cohort, while the model for Task 2 was trained and validated exclusively on the intersection cohort to ensure that all samples had the requisite imaging and molecular information. The final distribution of patients across these cohorts is summarized in Table~\ref{tab:brats_dist}. 
\begin{table}[!t] 
\centering \caption{Distribution of BraTS 2021 dataset cohorts used in this study.} 
\label{tab:brats_dist} 
\begin{tabular}{@{}p{3.2cm}p{1.2cm}p{3.8cm}@{}}
\toprule \textbf{Dataset Cohort} & \textbf{\# Patients} & \textbf{Label Information} \\
\midrule Task 1: Segmentation & $1470$ & Pixel-wise tumor sub-region masks (NET/NCR, ED, ET) \\ 
Task 2: MGMT Classification & $672$ & Binary MGMT promoter methylation status \\ Intersection (Task 1 $\cap$ Task 2) & $663$ & Both Segmentation and MGMT labels \\ \bottomrule 
\end{tabular} 
\end{table} 
The curated labels and co-registered imaging volumes were organized hierarchically within our project codebase to facilitate efficient data loading and model training.

These labels and corresponding imaging volumes are organized within our codebase as \texttt{BraTS2021\_train\_labels.csv} and \texttt{BraTS2021\_test\_labels.csv}, with cached volumes and masks stored under \texttt{Dataset/Train/\$ID}.

\subsubsection{UCSD-PTGBM: External Validation Cohort} The UCSD-PTGBM dataset \cite{dataset2} was employed for the external validation of our model, which was trained on the RSNA-MICCAI BraTS 2021 dataset. This specific dataset was selected due to its recent release in 2025 and its direct relevance to our research objectives.

%\textbf{Dataset Curation and Harmonization:} 
\paragraph{Dataset Curation and Harmonization} The initial curation of the UCSD-PTGBM cohort involved reducing the original metadata provided by TCIA to retain only the patient identifier (UCSD\_ID) and the MGMT promoter methylation status. The MGMT labels were subsequently standardized into a binary scheme, where methylated cases were encoded as $1$ and unmethylated cases as $0$. Entries with missing or invalid values were systematically excluded. A subsequent {\em consistency check} was performed to ensure complete alignment between the label file and the imaging data. This process involved the removal of extraneous imaging folders lacking corresponding labels and the exclusion of label entries for which no valid imaging data existed. This rigorous harmonization resulted in a final, curated dataset of $109$ patients, each possessing both valid imaging data and {MGMT annotations}, thereby establishing a reliable external test set for model evaluation. 

%\textbf{Imaging Modalities:}
\paragraph{Imaging Modalities} To maintain consistency with the modalities present in the {BraTS 2021 training dataset}, only the four conventional MRI modalities from the UCSD-PTGBM cohort were utilized: T1-weighted pre-contrast (T1pre), T1-weighted post-contrast (T1post), T2-weighted (T2), and Fluid Attenuated Inversion Recovery (FLAIR). The provided tumor segmentation masks were also included. Additional advanced modalities within the UCSD-PTGBM dataset, such as ADC, ASL, DSC, RSI, and SWAN, were not incorporated into the present study but may be investigated in future research to evaluate their potential contribution to MGMT promoter methylation prediction.

%\textbf{Cohort Composition and Significance:}
\paragraph{Cohort Composition and Significance} The final harmonized cohort consisted of $109$ patients, balanced between $54$ methylated and $55$ unmethylated cases. The use of an external validation cohort, such as UCSD-PTGBM, is critical for addressing {\em domain shift}, a well-recognized challenge that significantly impedes the clinical translation of computational models \cite{ref15}.

\subsection{Preprocessing Pipeline}\label{sec:preprocessing}
\begin{figure*}[!t]
  \centering
  \includegraphics[width=\textwidth,keepaspectratio]{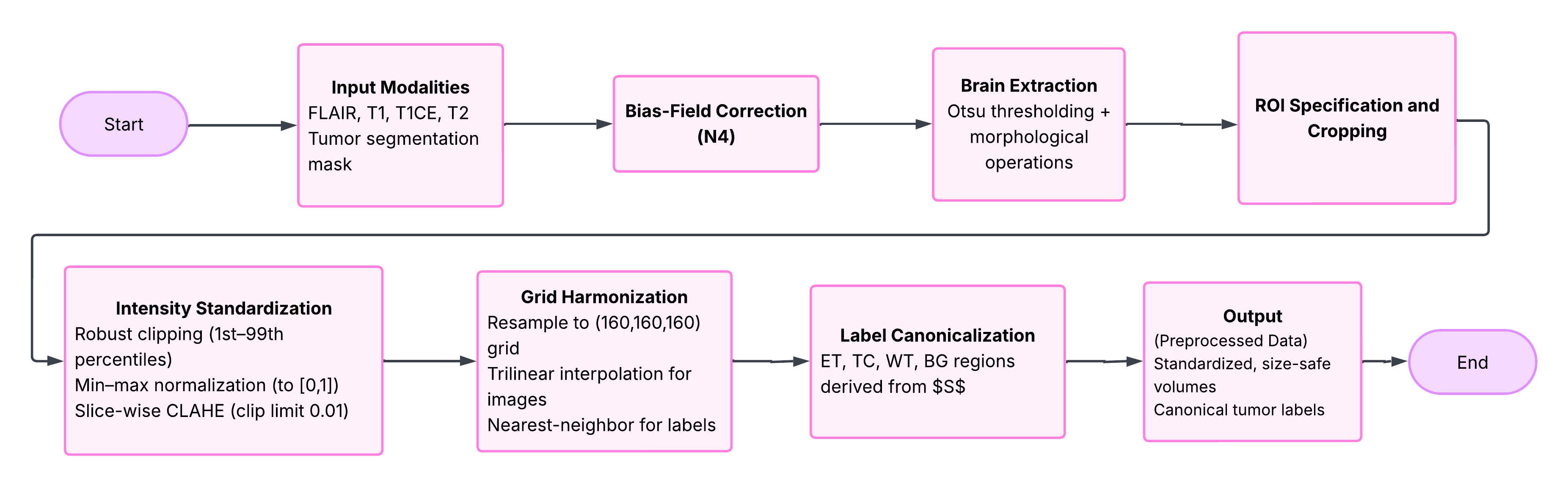}
  \caption{Overview of the preprocessing pipeline. The pipeline includes bias-field correction, brain extraction, ROI specification and cropping, intensity standardization, grid harmonization, and label canonicalization, resulting in standardized, size-safe volumes and canonical tumor labels.}
  \label{fig:preprocessing_pipeline}
\end{figure*}

A uniform, size-safe preprocessing pipeline was applied across both cohorts:
\emph{(i)} RSNA--MICCAI BraTS~2021 (FLAIR, T1, T1CE, T2 with provided tumor segmentation and MGMT labels), and
\emph{(ii)} UCSD\_PTGBM (FLAIR, T1pre, T1post, T2 with provided tumor mask and MGMT labels).
The pipeline corrects low-frequency bias fields, removes non-brain tissue, standardizes intensities within the region of interest (ROI), harmonizes voxel grids, and ensures consistent volumes for training, validation, and external testing. The overall preprocessing workflow is summarized in Figure~\ref{fig:preprocessing_pipeline}.

\subsubsection{Notation}
For subject $s$, the four modalities are denoted
\[
\mathcal{V}_s=\big\{V^{\text{FLAIR}},\,V^{\text{T1}},\,V^{\text{T1CE}},\,V^{\text{T2}}\big\}\in\mathbb{R}^{D\times H\times W},
\]
with voxel order $(z,y,x)$. Tumor labels follow the BraTS convention $S\in\{0,1,2,4\}^{D\times H\times W}$, encoding background~(0), NCR/NET~(1), ED~(2), and ET~(4). MGMT status is binary $y_s\in\{0,1\}$.

\subsubsection{Bias-Field Correction (N4)}
Observed images are modeled as
\[
V_{\text{obs}}=V_{\text{true}}\cdot B,
\]
or additively in the log-domain:
\[
\log V_{\text{obs}}=\log V_{\text{true}}+\log B.
\]
N4 estimates a smooth $\widehat{\log B}$ and produces
\[
V_{\text{corr}}=\frac{V_{\text{obs}}}{\exp\!\big(\widehat{\log B}\big)}.
\]
To preserve geometric fidelity, the estimated field is resampled back to the original grid before correction.

\subsubsection{Brain Extraction}
As segmentation masks were available for both cohorts, the brain region was localized using Otsu thresholding with hole filling and morphological closing. The resulting mask $M^{\text{brain}}\in\{0,1\}^{D\times H\times W}$ was applied as
\[
V_{\text{brain}}=V_{\text{corr}}\odot M^{\text{brain}}.
\]

\subsubsection{ROI Specification and Cropping}
The ROI $R$ was defined as tumor tissue ($R=\mathbf{1}\{S>0\}$). A minimal bounding box $\mathcal{B}(R)$ was computed and padded by 12~voxels per dimension. All modalities were cropped to this box to confine subsequent operations to the relevant anatomy.

\subsubsection{Intensity Standardization in the ROI}
All operations were restricted to the effective ROI $\widetilde{R}$:
\[
\widetilde{R}=
\begin{cases}
\{\text{finite voxels}\}, & |R|<N_{\min},\\
R\cap\{\text{finite voxels}\}, & \text{otherwise},
\end{cases}
\quad N_{\min}=64.
\]
For each modality $X$:
\begin{enumerate}
  \item \textbf{Robust clipping} to the 1st and 99th percentiles inside $\widetilde{R}$.
  \item \textbf{Min--max normalization}:
  \[
  X\leftarrow \frac{X-p_1}{\max(p_{99}-p_1,\varepsilon)},\qquad \varepsilon=10^{-8},
  \]
  with global clamping to $[0,1]$.
  \item \textbf{Slice-wise CLAHE} on axial slices within $\widetilde{R}$ using a clip limit of 0.01, skipping near-constant slices.
\end{enumerate}

\subsubsection{Grid Harmonization}
Volumes were resampled to a fixed grid $(D',H',W')=(160,160,160)$ for batched 3D processing. Given original spacing $\Delta=(\Delta_x,\Delta_y,\Delta_z)$ and size $N=(W,H,D)$, the target spacing was
\[
\Delta'=\Delta\odot\frac{N}{N'},
\]
with $N'=(W',H',D')$. Trilinear interpolation was used for images and nearest neighbor for labels.

\subsubsection{Label Canonicalization}
From $S\in\{0,1,2,4\}$, mutually exclusive maps were derived:
\begin{multline}
\text{ET}=\mathbf{1}\{S=4\}, \quad
\text{TC}=\mathbf{1}\{S=1\vee S=4\}, \\
\text{WT}=\mathbf{1}\{S=1\vee S=2\vee S=4\}, \quad
\text{BG}=\mathbf{1}\{S=0\}.
\end{multline}

\subsubsection{Design Rationale}
\begin{itemize}
  \item ROI-restricted statistics mitigate background-driven variability and stabilize intensity scales across subjects.
  \item Slice-wise CLAHE enhances local contrast while avoiding noise amplification in uniform slices.
  \item Size-safe N4 correction avoids geometric drift by correcting on the native grid.
  \item The fixed $160^3$ grid balances anatomical coverage with GPU memory efficiency.
\end{itemize}

\subsubsection{Default Parameterization}
\begin{figure}[!t]
  \centering
  \includegraphics[width=\linewidth]{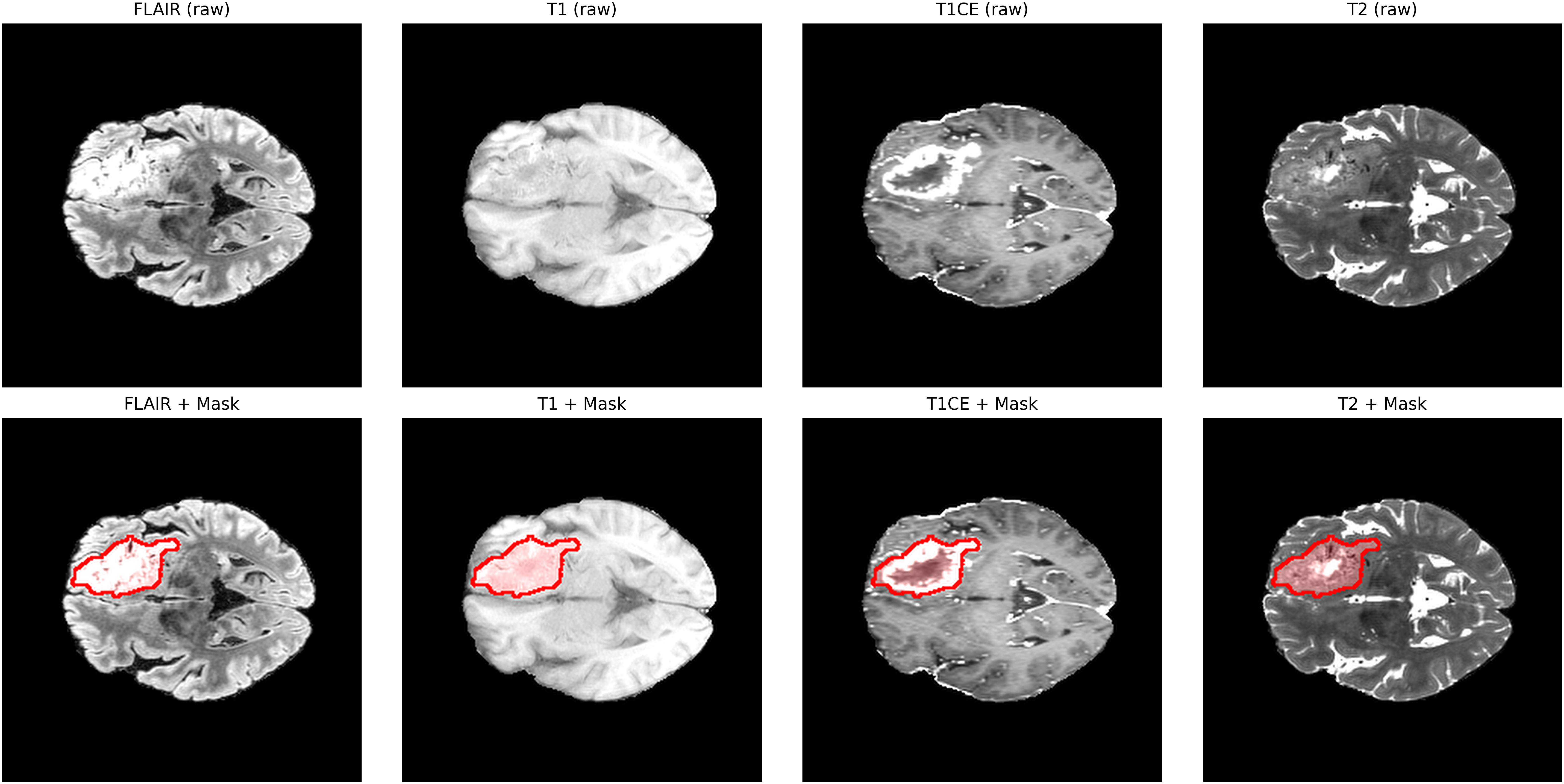}
  \caption{Representative axial slices for one subject. Top: raw FLAIR/T1/T1CE/T2. Bottom: normalized images with tumor overlay; all modalities use the same slice index.}
  \label{fig:preproc_panel}
\end{figure}

The preprocessing pipeline employed the following default parameters: padding of 12~voxels in each dimension, resampling to a target volume size of $(160,160,160)$, intensity clipping to the 1st and 99th percentiles, slice-wise CLAHE with a clip limit of 0.01, and an ROI size threshold of $N_{\min}=64$ voxels.
A qualitative example of raw versus normalized slices, with tumor overlay, is shown in Fig.~\ref{fig:preproc_panel}.

\subsection{Brain Tumor Segmentation}
\label{sec:segmentation}
We employ a three-dimensional residual U-Net backbone augmented with Variable-Scale Attention (\textbf{ResUNetVSA}) for multi-modal glioma segmentation. The design emphasizes (i) scale-adaptive feature enrichment via depthwise-dilated attention, (ii) channel- and spatial-wise gating to prioritize tumor-relevant structures, and (iii) a hybrid training objective coupling regional fidelity (cross-entropy, soft Dice) with a boundary-sensitive regularizer.

\subsubsection{Input from Preprocessing Cache}
Following the preprocessing pipeline, each subject is represented by standardized, resampled, and harmonized tensors. The four MRI modalities \{FLAIR, T1, T1CE, T2\} are stacked into a volume of shape $[4,160,160,160]$, while tumor labels are stored as a one-hot tensor of the same shape, encoding \{\textsf{BG}, \textsf{ET}, \textsf{TC\_only}, \textsf{WT\_only}\}. For segmentation experiments, these cached tensors are accessed directly through the \texttt{BratsCacheSegDataset} loader, eliminating repeated parsing of raw NIfTI files and ensuring reproducibility across training, validation, and testing.

\subsubsection{Network Architecture: ResUNetVSA}
The backbone follows an encoder–decoder paradigm with skip connections. Each stage integrates residual convolutional blocks with a variable-scale attention (VSA) module:
\begin{itemize}
  \item \textbf{Residual blocks.} Two group-normalized $3\times3\times3$ convolutions with SiLU activations and an identity skip connection ensure stable optimization at 3D resolution.
  \item \textbf{Variable-Scale Attention (VSA).} For input feature map $F\in\mathbb{R}^{C\times D\times H\times W}$, multiple depthwise convolutions with dilations $d\in\{1,2,3/4\}$ capture short- to mid-range context. Their outputs are concatenated and fused by a pointwise convolution, then gated by:
  \begin{enumerate}
    \item \textit{Channel gating (SE)}:  
    \(
    \mathrm{SE}(F)=F\cdot\sigma(W_2\,\delta(W_1\,\mathrm{GAP}(F))),
    \)
    where $\mathrm{GAP}$ is global average pooling, $\delta$ is SiLU, and $\sigma$ is sigmoid.
    \item \textit{Spatial gating}: a one-channel attention mask derived from per-voxel mean and max responses, followed by a $k^3$ convolution and sigmoid activation.
  \end{enumerate}
  This composite VSA enriches features at multiple scales while remaining computationally efficient.
  \item \textbf{Down/Up-sampling.} Max-pooling reduces spatial resolution in the encoder; transposed convolutions restore resolution in the decoder. Skip connections concatenate encoder and decoder features at corresponding scales.
  \item \textbf{Output head.} A $1\times1\times1$ convolution projects to $C_{\text{seg}}=4$ logits, followed by softmax normalization.
\end{itemize}

\subsubsection{Hybrid Objective with Boundary Regularization}
Let $Z\in\mathbb{R}^{4\times D\times H\times W}$ be the logits and $P=\mathrm{softmax}(Z)$ the predicted probabilities. With $Y$ as the one-hot ground truth, the segmentation loss is
\begin{multline}
\mathcal{L}_{\mathrm{seg}}(Z,Y; e)
= \underbrace{\mathrm{CE}\!\big(Z,\arg\max\nolimits_{c} Y_c\big)}_{\text{voxel-wise fidelity}} 
+ \underbrace{\mathcal{L}_{\mathrm{Dice}}(P,Y)}_{\text{overlap consistency}} \\
+ \lambda(e)\,\underbrace{\mathcal{R}_{\partial \Omega}(Y)}_{\text{boundary regularizer}},
\label{eq:seg_loss}
\end{multline}
where $e$ is the epoch index.

\paragraph{Soft Dice loss (foreground classes).}
For $c\in\{\textsf{ET},\textsf{TC\_only},\textsf{WT\_only}\}$,
\begin{equation}
\mathcal{L}_{\mathrm{Dice}}
= 1 - \frac{1}{3}\sum_{c=1}^{3}\frac{2\sum P_c Y_c + \varepsilon}{\sum P_c + \sum Y_c + \varepsilon},
\qquad \varepsilon=10^{-6}.
\end{equation}

\paragraph{Cosine-ramped boundary regularizer.}
Let $Y_{\mathrm{fg}}=\mathbf{1}\{\sum_{c=1}^{3}Y_c>0\}$. The morphological gradient is approximated as
\(
G = \mathrm{dilate}(Y_{\mathrm{fg}})-\mathrm{erode}(Y_{\mathrm{fg}}).
\)
We penalize mean edge activation as
\(
\mathcal{R}_{\partial \Omega}(Y) = \tfrac{1}{DHW}\sum G.
\)
Its weight follows a cosine ramp:
\begin{equation}
\lambda(e) = \lambda_{\max}\,\frac{1-\cos\big(\pi e/E\big)}{2},
\label{eq:ramp}
\end{equation}
where $E$ is the total epochs and $\lambda_{\max}$ (set to 0.3) controls regularization strength. This emphasizes region learning in early stages and progressively sharpens boundaries.

\subsubsection{Training Protocol}
Optimization uses AdamW (learning rate $10^{-3}$, weight decay $10^{-4}$) with mixed precision, gradient clipping at $1.0$, and a \texttt{ReduceLROnPlateau} scheduler on validation Dice. Data augmentation includes 3D flips, small-angle rotations, elastic and anisotropic warps, and mild intensity perturbations. Gradient accumulation supports effective batch sizing under GPU memory constraints. Early stopping monitors the mean Dice across \{\textsf{ET}, \textsf{TC}, \textsf{WT}\}.

\subsubsection{Evaluation Metric}
Predicted labels $\hat{L}=\arg\max_c P_c$ are compared with ground truth $L=\arg\max_c Y_c$. For each region $r\in\{\textsf{ET},\textsf{TC},\textsf{WT}\}$,
\[
\mathrm{Dice}_r=\frac{2|\hat{r}\cap r|}{|\hat{r}|+|r|+\varepsilon}.
\]
We report Dice scores for ET, TC, WT, and their macro average.

\subsubsection{Implementation Notes}
\begin{figure*}[!thb]
\centering
\includegraphics[width=.85\linewidth]{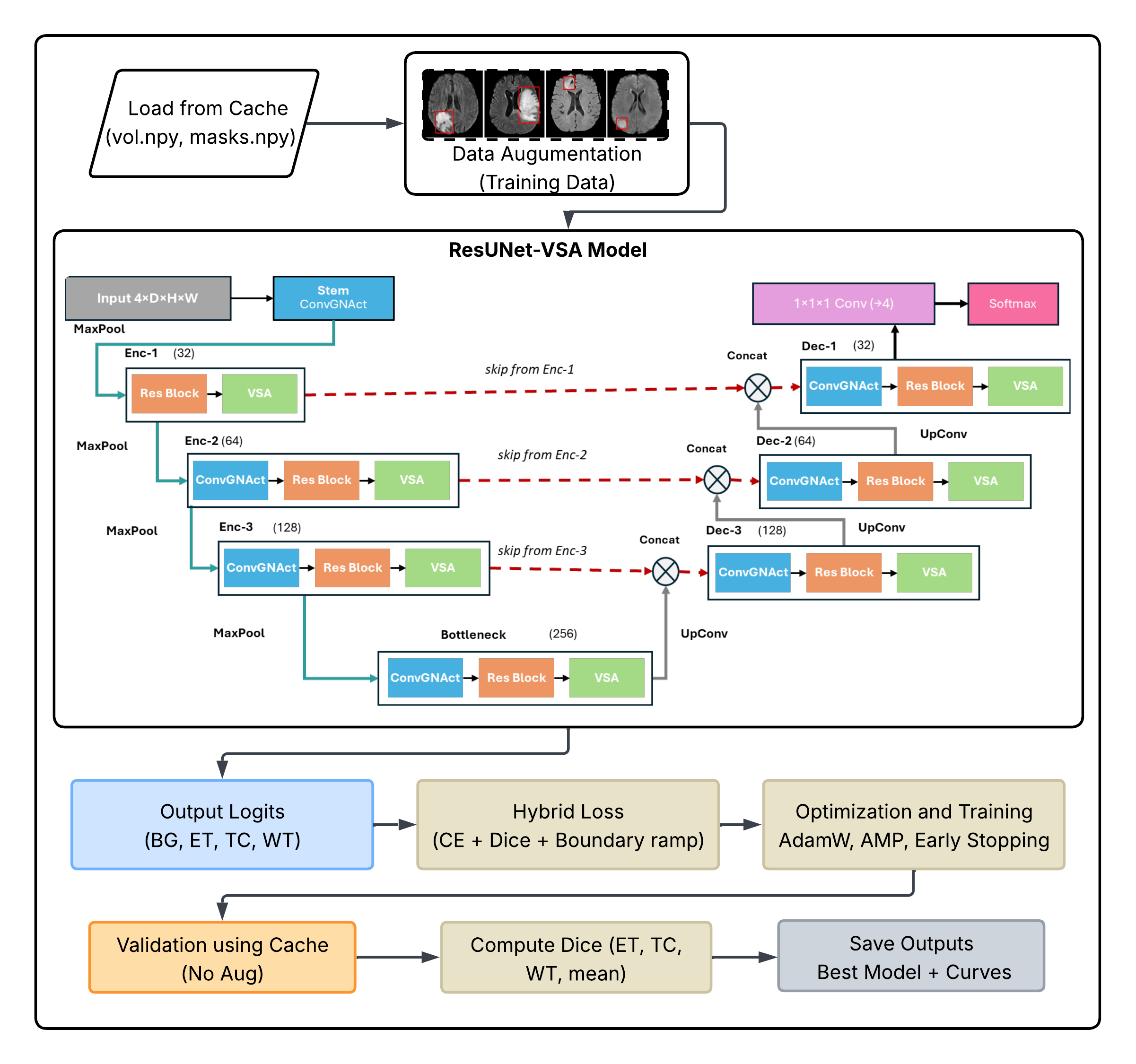}
\caption{Segmentation pipeline. The ResUNetVSA backbone integrates residual convolutional blocks, variable-scale attention (VSA), channel and spatial gating, and skip connections. Training employs a hybrid loss combining cross-entropy, soft Dice over tumor subregions, and a cosine-ramped boundary regularizer to improve boundary delineation.}
\label{fig:seg_flow}
\end{figure*}

The model is implemented in PyTorch with multi-GPU training via DistributedDataParallel. Group normalization stabilizes small-batch 3D training. SE and spatial gates add minimal overhead while enhancing attention to rim and peritumoral edema. Variable dilations in VSA capture multi-scale context without excessive depth or parameters. Brain Tumor Segmentation pipeline is shown in \ref{fig:seg_flow}.

\subsection{Segmentation-Probability Channels (ET/TC/WT)}
From the trained segmentation network, we compute per-voxel class probabilities \(P(c=k \mid \mathbf{x})\) for \(k\in\{0,1,2,3\}\) and derive three probability channels appended to the classifier input:
\begin{itemize}[leftmargin=*, itemsep=2pt]
    \item \textbf{ET:} \(\mathrm{ET} = P(c=3)\) \;(Enhancing Tumor).
    \item \textbf{TC:} \(\mathrm{TC} = P(c=1) + P(c=3)\) \;(Tumor Core).
    \item \textbf{WT:} \(\mathrm{WT} = 1 - P(c=0)\) \;(Whole Tumor).
\end{itemize}

\subsection{Radiomics Extraction and Normalization}
PyRadiomics is applied within the binarized tumor on T1CE (first-order, GLCM, GLRLM, GLSZM, GLDM, NGTD, shape; \texttt{binWidth}=5; \texttt{resampledPixelSpacing}=[2.0,2.0,2.0]; BSpline). Features are standardized using training-set mean/std and these statistics are reused at validation/inference (including UCSD-PTGBM).

\subsection{MGMT Classification}

\begin{figure}[t]
    \centering
    \includegraphics[width=1\linewidth]{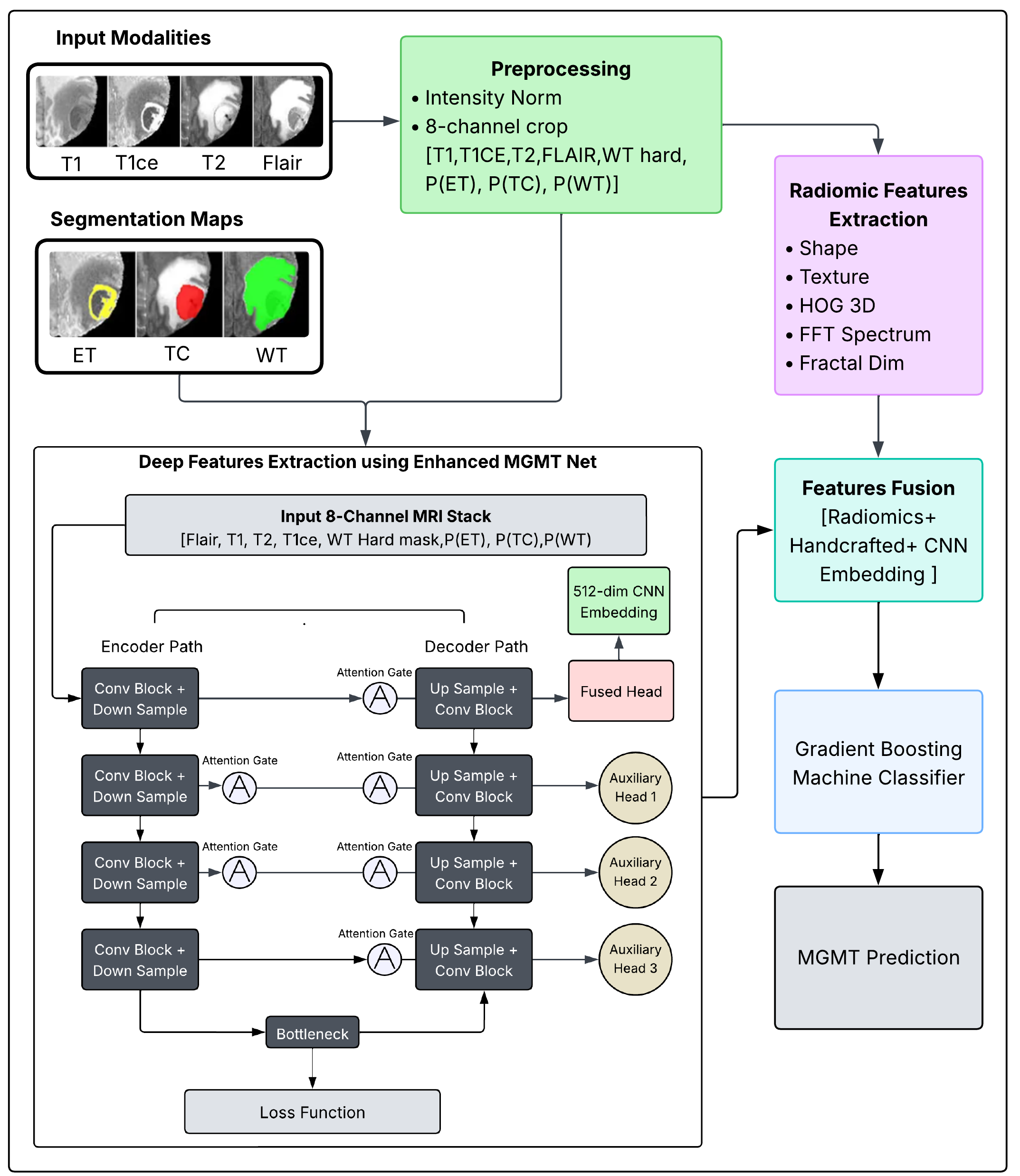}
    \caption{Overview of the proposed hybrid pipeline for MGMT promoter methylation prediction. Multi-parametric MRIs and segmentation masks are preprocessed into 8-channel crops, processed through radiomics and deep learning feature extraction branches, fused into a joint representation, and classified with a gradient boosted machine (GBM).}
    \label{fig:mgmt_pipeline}
\end{figure}

For MGMT promoter methylation prediction, we designed a \textbf{hybrid framework} that integrates handcrafted radiomic descriptors with deep volumetric features learned from an 8-channel convolutional encoder (Fig.~\ref{fig:mgmt_pipeline}).  

\subsubsection{Feature representation.}  
From each patient’s multi-parametric MRI (T1, T1CE, T2, and FLAIR) and tumor segmentation, we constructed an 8-channel input tensor by appending a hard whole tumor mask $WT$ together with three synthesized soft probability maps:
\[
[\text{T1}, \text{T1CE}, \text{T2}, \text{FLAIR}, WT, P(ET), P(TC), P(WT)],
\]
where $P(ET)$, $P(TC)$, and $P(WT)$ are Gaussian-smoothed, nested probability maps satisfying $ET \subset TC \subset WT$.  

Two complementary feature sets were then extracted:  
\begin{itemize}
    \item \textit{Radiomic features}: shape, first-order, and texture descriptors (GLCM, GLRLM, GLSZM, GLDM, NGTDM) computed with \texttt{PyRadiomics}, together with handcrafted descriptors such as 3D histogram-of-oriented-gradients (HOG3D), FFT-based spectral statistics, histogram moments, and fractal dimension.
    \item \textit{Deep features}: an enhanced 3D encoder network (\texttt{EnhancedMGMTNet}) operating on the 8-channel input. The network employs multi-scale encoding with attention-gated skip connections and produces three auxiliary classification heads $(h_1,h_2,h_3)$ as well as a fused head. The penultimate fused representation is a compact 512-dimensional embedding.
\end{itemize}

\subsubsection{Optimization of CNN.}  
CNN heads were jointly trained using a composite ensemble loss:
\begin{equation}
\mathcal{L}_{\text{ens}} = \sum_{h \in \{h_1,h_2,h_3,h_f\}} \text{BCE}(h,y) 
+ \lambda_{\text{var}} \,\text{Var}(p_h) 
+ \lambda_{\text{jsd}} \,\text{JSD}(p_h),
\end{equation}
where $\text{BCE}$ is the binary cross-entropy, $\text{Var}(p_h)$ penalizes disagreement across head probabilities, and $\text{JSD}(p_h)$ is the Jensen–Shannon divergence regularizer. Temperature scaling was applied post hoc to calibrate fused logits.  

\subsubsection{Feature fusion.}  
The final feature vector was constructed by concatenating CNN embeddings with radiomic and handcrafted descriptors:
\begin{equation}
    \mathbf{f} = \big[ \mathbf{f}_{\text{cnn}} \;\oplus\; \mathbf{f}_{\text{rad}} \;\oplus\; \mathbf{f}_{\text{extras}} \big].
\end{equation}

\subsubsection{Classifier.}  
Instead of a linear head, we employed gradient boosted machines (XGBoost / CatBoost) trained on $\mathbf{f}$ to obtain the MGMT methylation probability:
\begin{equation}
    p(y=1|\mathbf{f}) = \Phi_{\text{GBM}}(\mathbf{f}),
\end{equation}
where $\Phi_{\text{GBM}}$ denotes the nonlinear decision function of the boosted ensemble.  

\subsubsection{Evaluation.}  
Model parameters were optimized with AdamW and mixed-precision training. Radiomics features were standardized prior to GBM training. Performance was evaluated on the validation split using ROC--AUC as the primary metric, with accuracy, F1-score, and Brier score reported as secondary measures.  

\subsection{Training/Validation Protocol and External Test}
Segmentation is trained on BraTS~2021 Task~1; MGMT classification on the Task~1$\cap$Task~2 intersection (663 subjects). We use stratified 80/20 splits (different seeds for segmentation and classification). All tensors are cached; radiomics normalization stats are stored per run. For \textbf{UCSD-PTGBM} external testing, we freeze both models, apply the same preprocessing, generate masks and ET/TC/WT maps with the segmenter, and run MGMT inference with the learned decision threshold -- no fine-tuning.

\subsection{Evaluation Metrics}

The proposed framework was evaluated using complementary metrics for segmentation and classification tasks.  

\subsubsection{Segmentation Metrics.}  
For tumor subregion segmentation (enhancing tumor: ET, tumor core: TC, and whole tumor: WT), two metrics were used:

\paragraph{Dice Similarity Coefficient (DSC).}  
The DSC measures volumetric overlap between prediction $\hat{S}$ and reference $S$ \cite{evalref1}:  
\begin{equation}
\text{DSC}(S, \hat{S}) = \frac{2|S \cap \hat{S}|}{|S| + |\hat{S}|},
\end{equation}
where $|S|$ and $|\hat{S}|$ denote the number of voxels in the ground-truth and predicted masks, respectively, and $|S \cap \hat{S}|$ is their intersection. A macro-DSC was reported as the arithmetic mean over ET, TC, and WT.  

\paragraph{95th percentile Hausdorff distance (HD95).}  
The HD95 quantifies boundary similarity by computing the 95th percentile of bidirectional surface distances \cite{evalref2}:  
\begin{equation}
\text{HD}_{95}(S, \hat{S}) = \max \Bigg\{
\underset{x \in S}{\operatorname{quantile}_{95}} \, d(x, \hat{S}), \;
\underset{y \in \hat{S}}{\operatorname{quantile}_{95}} \, d(y, S)
\Bigg\},
\end{equation}
where $d(a,B)$ is the minimum Euclidean distance from point $a$ to set $B$.

\subsubsection{Classification Metrics.}  
For MGMT promoter methylation classification, the following metrics were adopted:

\paragraph{Receiver Operating Characteristic Area Under the Curve (ROC--AUC).}  
The ROC curve plots true positive rate (TPR, sensitivity) against false positive rate (FPR) across thresholds. The AUC is the integral under this curve:  
\begin{equation}
\text{AUC} = \int_{0}^{1} \text{TPR}(\text{FPR}) \, d(\text{FPR}).
\end{equation}

\paragraph{True Positive Rate (TPR, Sensitivity).}  
\begin{equation}
\text{TPR} = \frac{\text{TP}}{\text{TP} + \text{FN}},
\end{equation}
where TP are correctly predicted methylated cases and FN are methylated cases misclassified as unmethylated.  

\paragraph{True Negative Rate (TNR, Specificity).}  
\begin{equation}
\text{TNR} = \frac{\text{TN}}{\text{TN} + \text{FP}},
\end{equation}
where TN are correctly predicted unmethylated cases and FP are unmethylated cases misclassified as methylated.  

\paragraph{False Positive Rate (FPR).}  
\begin{equation}
\text{FPR} = \frac{\text{FP}}{\text{FP} + \text{TN}}.
\end{equation}

\paragraph{Classification Accuracy (ACC).}  
\begin{equation}
\text{ACC} = \frac{\text{TP} + \text{TN}}{\text{TP} + \text{TN} + \text{FP} + \text{FN}}.
\end{equation}

\paragraph{Precision and Recall.}  
\begin{equation}
\text{Precision} = \frac{\text{TP}}{\text{TP} + \text{FP}}, \qquad
\text{Recall} = \frac{\text{TP}}{\text{TP} + \text{FN}}.
\end{equation}

\paragraph{Macro F1-score.}  
\begin{equation}
\text{F1}_{\text{macro}} = \frac{1}{2} \sum_{c=0}^1 
\frac{2 \cdot \text{Precision}_c \cdot \text{Recall}_c}{\text{Precision}_c + \text{Recall}_c},
\end{equation}
averaged across both classes $c \in \{0,1\}$.

\paragraph{Brier Score.}  
The Brier score measures calibration of predicted probabilities $\hat{p}_i$:  
\begin{equation}
\text{Brier} = \frac{1}{N} \sum_{i=1}^N (\hat{p}_i - y_i)^2,
\end{equation}
where $y_i \in \{0,1\}$ is the ground-truth label.

\paragraph{Confusion Matrix.}  
A confusion matrix was constructed to summarize counts of true positives (TP), true negatives (TN), false positives (FP), and false negatives (FN), enabling visualization of class-wise performance.

\section{Results and Discussion}
\label{sec:results}
This section reports: (i) tumor segmentation quality on BraTS~2021, (ii) MGMT classification under cross-validation on the BraTS~2021 intersection cohort, (iii) external validation on UCSD-PTGBM, (iv) ablation of fusion strategies, (v) calibration and reliability analysis, (vi) explainability outcomes via Grad-CAM/SHAP, and (vii) a brief evaluation of our GUI application.

\subsection{Experimental Setup}
All experiments were conducted on the University of Idaho HPC cluster. 
Training and inference ran on nodes equipped with NVIDIA RTX A6000 GPUs (48 GB VRAM) and up to 64 CPU cores with 220 GB RAM. For additional cross-validation runs, NVIDIA A100 (80 GB) and RTX 4090 (24 GB) GPUs were also employed.

The software environment consisted of Python 3.8.11, PyTorch 2.1.2 with CUDA 11.8, and MONAI 1.3 for medical imaging pipelines. Radiomics were extracted using PyRadiomics 3.0. Segmentation preprocessing used SimpleITK (N4 bias correction) and scikit-image (CLAHE). Explainability modules leveraged Captum (for SHAP and Grad-CAM) and custom visualization code integrated into our GUI application.

Training hyperparameters were as follows: batch size 2, initial learning rate $3\times10^{-4}$ with cosine annealing warm restarts, and early stopping based on validation loss. For MGMT classification, we 
trained using focal loss ($\gamma=2$) with class-balanced positive weights derived from the training split. Each experiment was repeated across five 
folds with stratified splits to ensure robustness.

\subsection{Segmentation Results}
\subsubsection{Evaluation Metrics}
We evaluate segmentation accuracy using the Dice similarity coefficient (Dice) and the 95th percentile Hausdorff distance (HD95), computed on a per-case basis and then aggregated. The Dice score measures overlap between the predicted tumor region $\hat{S}$ and the ground truth region $S$ as
$$ \text{Dice}(\hat{S}, S) \;=\; \frac{2\,|\hat{S} \cap S|}{\,|\hat{S}| + |S|\,}\,. $$,
ranging from 0 (no overlap) to 1 (perfect overlap). We report the micro-averaged Dice, meaning we compute the Dice over the total tumorous volume pooled across all patients (this effectively weights each case by its tumor size and gives a single overall overlap measure). The HD95 measures the border error: it is the $95^{\text{th}}$
percentile of the Hausdorff distance between the predicted and true tumor surfaces. In practice, we compute the distance (in mm) from each point on the prediction surface to the nearest point on the true surface and vice versa, take the maximum of these distances (the Hausdorff distance), and then report the 95th percentile (which excludes outlier points). A lower HD95 (in millimeters) indicates better spatial precision of the segmentation boundaries (with zero being a perfect alignment). Together, a high Dice and low HD95 signify that the model accurately captures the tumor’s volume and shape, which is crucial for reliable downstream radiomic feature extraction and MGMT methylation prediction. The described preprocessing and segmentation pipeline thus provides a robust basis for subsequent radiogenomic analysis, isolating the tumor region and ensuring consistent, high-quality input to the MGMT classification model.
\subsubsection{Optimization and Convergence}
\begin{figure}[htb!]
    \centering
    \includegraphics[width=\linewidth]{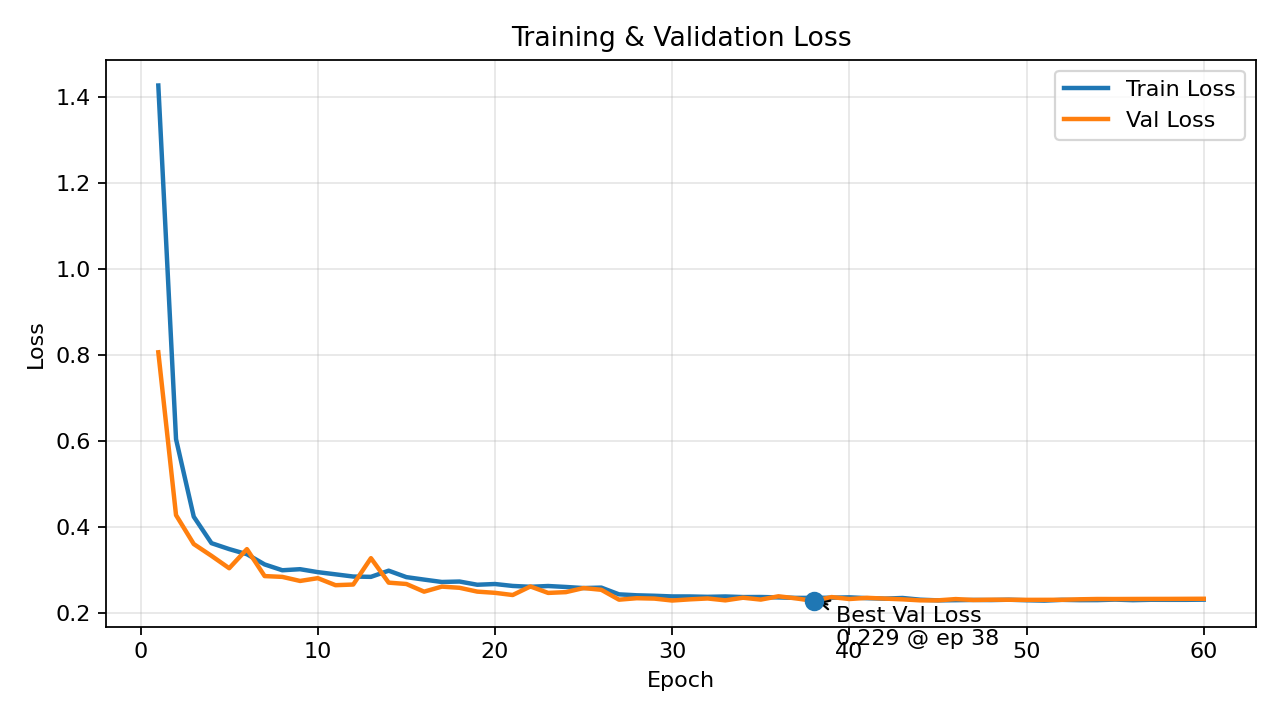}
    \caption{Training and validation loss versus epoch.}
    \label{fig:loss_curve}
\end{figure}

Training on the RSNA MICCAI BraTS 2021 dataset converged stably, with validation loss dropping from $0.81$ to $0.23$ by epoch $38$ (see Figure~\ref{fig:loss_curve}). Loss reductions coincided with scheduled learning-rate decays, and late epochs exhibited a stable loss plateau (mean $\approx 0.234$, epochs $50$–$60$). The mean epoch time was approximately $419$ s, giving a total training time of $7$ hours.

\subsubsection{Segmentation Quality (Dice Score)}

Overall segmentation quality, measured by the macro Dice coefficient (DSC), improved from $0.804$ in the initial epoch to a best of $0.913$ (epoch $52$), remaining above $0.91$ from epoch $27$ onwards (last-5-epoch mean $\sim0.912$), indicating strong generalization.

\begin{figure}[htb!]
    \centering
    \includegraphics[width=\linewidth]{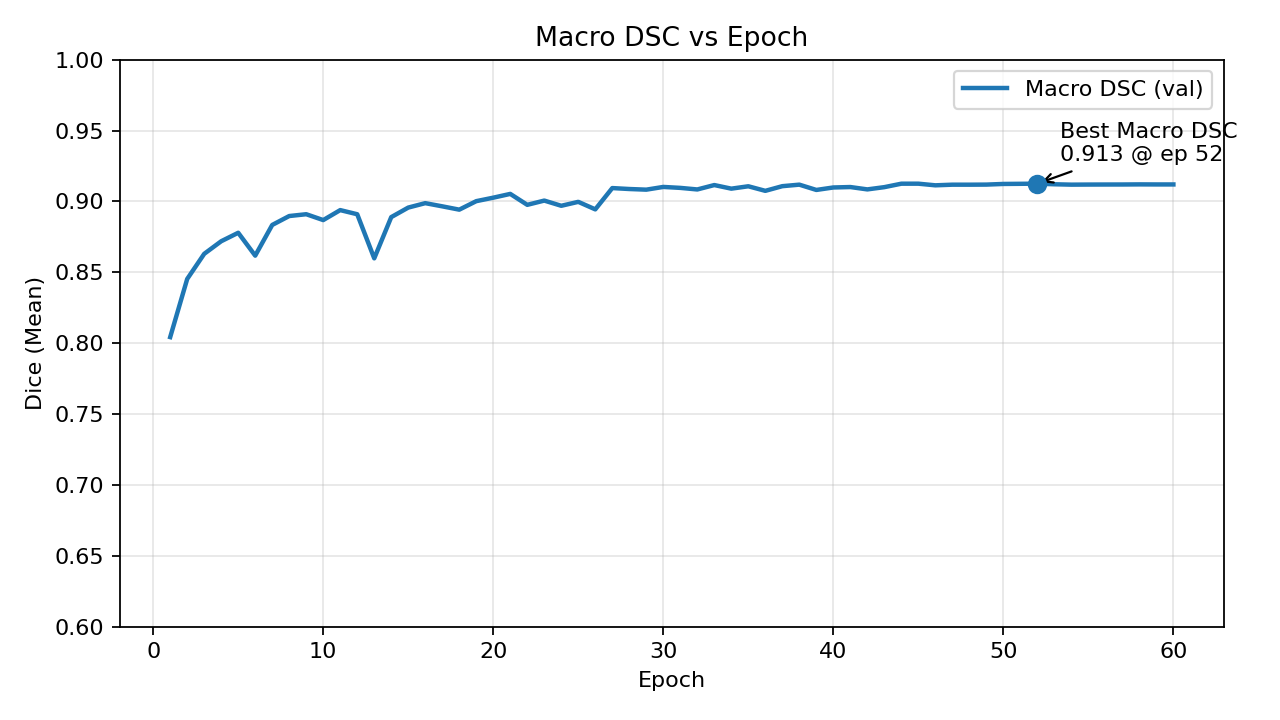}
    \caption{Macro Dice coefficient versus epoch.}
    \label{fig:macro_dice}
\end{figure}

\begin{figure}htb!]
    \centering
    \includegraphics[width=\linewidth]{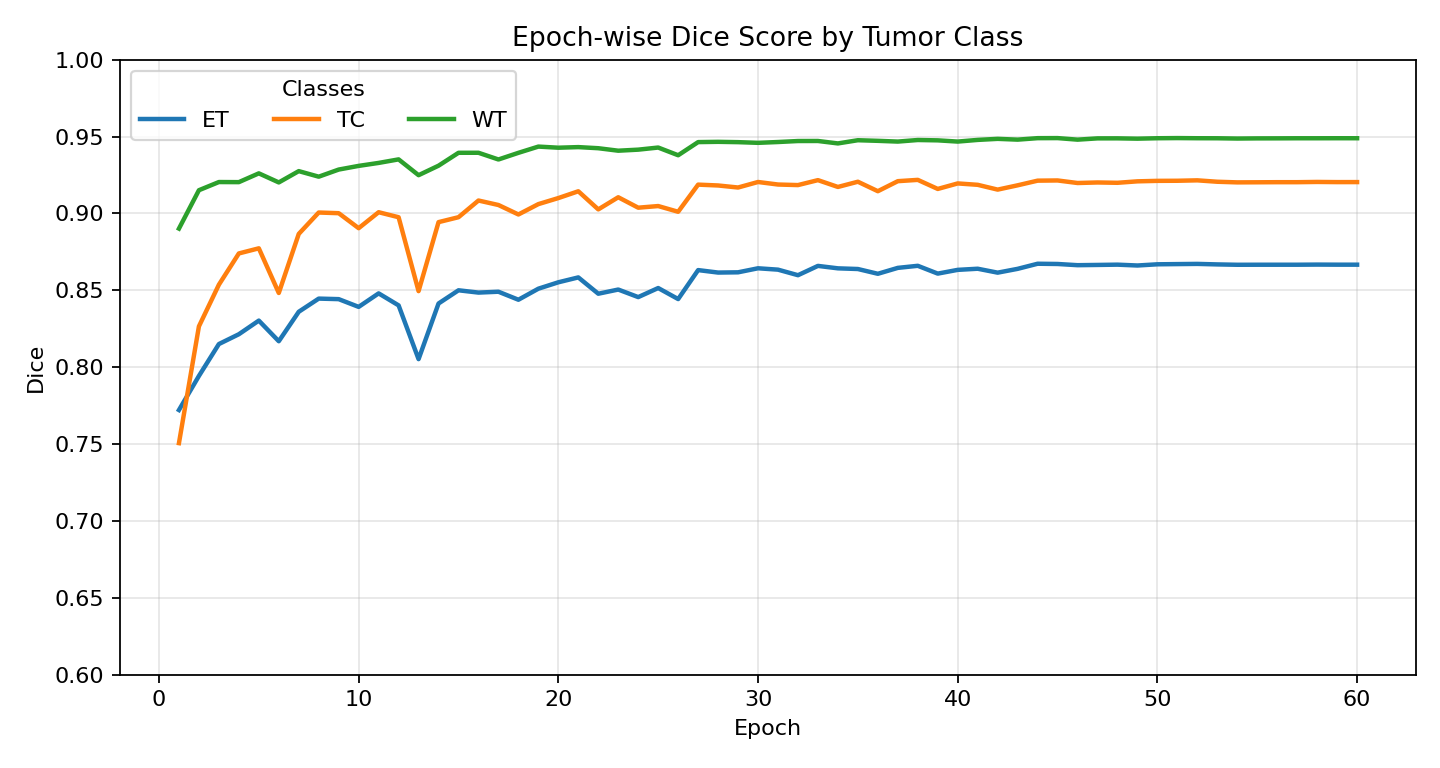}
    \caption{Dice scores for ET, TC, and WT tumor regions across epochs.}
    \label{fig:class_dice}
\end{figure}

Class-wise Dice scores (Figure~\ref{fig:class_dice}) show robust and balanced subregion segmentation:
\begin{itemize}
    \item \textbf{Enhancing Tumor (ET):} peaked at $0.867$ (epoch $44$), sustained above $0.863$ from epoch $37$ onward.
    \item \textbf{Tumor Core (TC):} peaked at $0.922$ (epoch $38$), stabilized at $\sim0.920$.
    \item \textbf{Whole Tumor (WT):} peaked at $0.949$ (epoch $51$), sustained at $\sim0.949$ in the last epochs.
\end{itemize}

\subsubsection{Hausdorff Distance (HD95)}
We further evaluated the segmentation quality using the 95$^{\text{th}}$ percentile Hausdorff Distance (HD95), which quantifies the boundary alignment between predicted and ground truth tumor regions. Lower HD95 values indicate better spatial agreement. Table~\ref{tab:hd95_results} reports the mean $\pm$ standard deviation (SD) and median values across the three tumor sub-regions. 
The following table summarizes the HD95 results across tumor subregions:

\begin{table}htb!]
\centering
\caption{HD95 results (in mm) for different tumor sub-regions.}
\label{tab:hd95_results}
\begin{tabular}{lcc}
\toprule
\textbf{Class} & \textbf{Mean $\pm$ SD} & \textbf{Median} \\
\midrule
Enhancing Tumor (ET) & 4.45 $\pm$ 0.41 & 4.48 \\
Tumor Core (TC)      & 4.70 $\pm$ 0.36 & 4.70 \\
Whole Tumor (WT)     & 1.65 $\pm$ 0.09 & 1.65 \\
\midrule
\textbf{Average}     & \textbf{3.60 $\pm$ 0.19} & \textbf{3.61} \\
\bottomrule
\end{tabular}
\end{table}

\subsection{Comparison with Prior Studies}

We compare our segmentation results to published models on BraTS 2021 below:

\begin{table}htb!]
    \centering
    \caption{Comparison of segmentation performance (Dice score) with leading methods on BraTS 2021}
    \label{tab:segmentation_comparison}
    \begin{tabular}{lcccc}
        \toprule
        \textbf{Ref. Study} & \textbf{WT} & \textbf{TC} & \textbf{ET} & \textbf{Macro Dice} \\
        \midrule
        Proposed (ResUNetVSA) & $0.949$ & $0.920$ & $0.867$ & $0.912$ \\
        \cite{segref1} & $0.919$ & $0.889$ & $0.889$ & $0.899$ \\
        \cite{segref2} & $0.927$ & $0.887$ & $0.860$ & $0.891$ \\
        \cite{segref3} & $0.913$ & $0.899$ & $0.848$ & $0.887$ \\
        \cite{ref15} & $0.84$ & $0.81$ & $0.80$ & $0.81$ \\
        \bottomrule
    \end{tabular}
\end{table}

Our model achieves competitive performance, matching or exceeding previously published scores for WT and TC regions, with robust ET segmentation.

\subsection{Radiomics Feature Distribution}
\begin{figure}htb!]
\centering
\includegraphics[width=\linewidth]{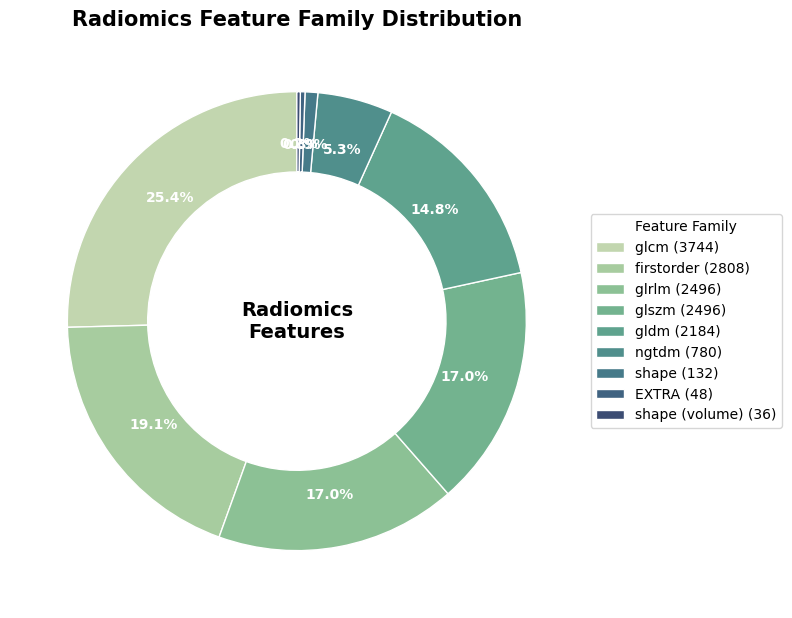}
\caption{Distribution of extracted radiomics feature families. 
Texture-based descriptors (\textit{GLCM}, \textit{GLRLM}, \textit{GLSZM}, and \textit{GLDM}) dominate the feature set, 
followed by first-order statistical features and shape-based metrics. 
EXTRA features correspond to histogram-based and fractal dimension statistics.}
\label{fig:feature_distribution}
\end{figure}

A total of radiomic features were extracted from all MRI modalities and tumor subregions using the PyRadiomics framework. 
Feature extraction for the entire BraTS 2021 cohort (577 subjects) required approximately 72~hours of computation, 
performed on NVIDIA RTX~A6000 GPUs (48~GB VRAM) with up to 64~CPU cores and 220~GB~RAM. 
To provide an overview of the feature composition, the extracted descriptors were categorized by their respective feature families 
and visualized as shown in Fig.~\ref{fig:feature_distribution}. 

As illustrated in Fig.~\ref{fig:feature_distribution}, 
the Gray Level Co-occurrence Matrix (\textit{GLCM}) features accounted for the largest proportion of the dataset (approximately 25\%), 
followed by First-order statistics (19\%), Gray Level Run Length Matrix (\textit{GLRLM}, 17\%), 
Gray Level Size Zone Matrix (\textit{GLSZM}, 17\%), and Gray Level Dependence Matrix (\textit{GLDM}, 15\%). 
Only a small fraction corresponded to Neighborhood Gray Tone Difference Matrix (\textit{NGTDM}), Shape, and EXTRA features.
The predominance of texture-related descriptors highlights their strong potential for capturing intra-tumoral heterogeneity 
and intensity non-uniformity, which are key imaging biomarkers in MGMT promoter methylation prediction.

\subsection{MGMT Classification: Cross-Validation on BraTS~2021}
Table~\ref{tab:cv_cls} summarizes five-fold cross-validation results on the BraTS~2021 intersection cohort. The proposed attention-gated radiomics fusion consistently outperforms both radiomics-only and deep-learning-only baselines across all metrics (primary metric: ROC--AUC; complementary: Accuracy and Macro-F1).

\begin{table}htb!]
\centering
\caption{Cross-validation performance on BraTS~2021 intersection cohort (5-fold CV). Mean~$\pm$~std across folds.}
\label{tab:cv_cls}
\resizebox{\columnwidth}{!}{
\begin{tabular}{lccc}
\toprule
\textbf{Method} & \textbf{ROC--AUC} & \textbf{Accuracy} & \textbf{Macro-F1} \\
\midrule
Radiomics-only (PyRadiomics + MLP) & 0.837 & 0.803 & 0.818 \\
Deep-only (3D ResNet-18 + ET/TC/WT) & 0.663 & 0.644 & 0.686 \\
\textbf{Fusion (ours)} & \textbf{0.871} & \textbf{0.866} & \textbf{0.857} \\
\bottomrule
\end{tabular}
}
\end{table}

\begin{figure}htb!]
\centering
\includegraphics[width=\linewidth]{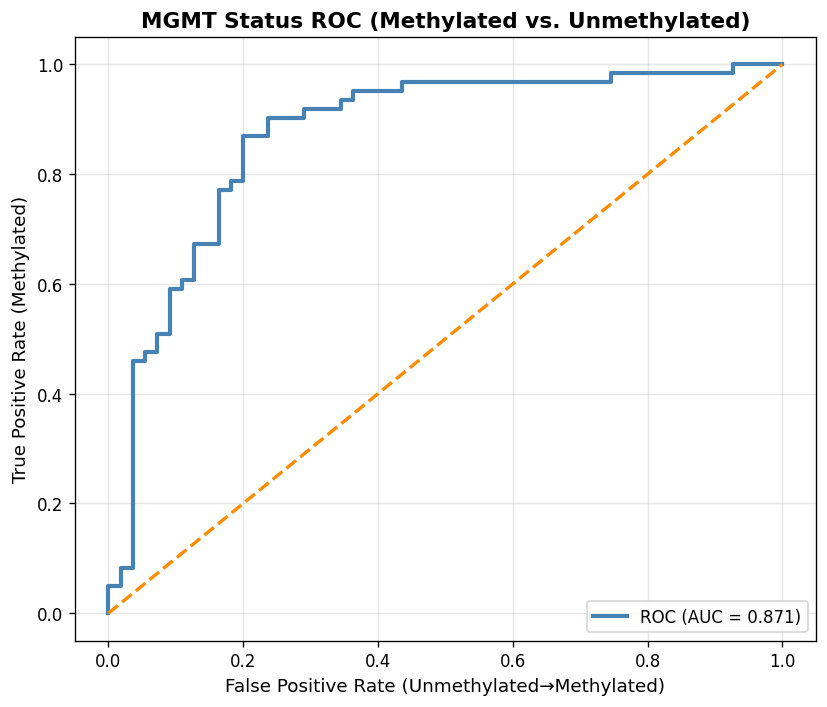}
\caption{ROC curve for five-fold cross-validation on the BraTS~2021 intersection cohort.}
\label{fig:cls_roc}
\end{figure}

Figure~\ref{fig:cls_roc} shows the corresponding ROC curve averaged across folds.

\begin{figure}htb!]
\centering
\includegraphics[width=.85\linewidth]{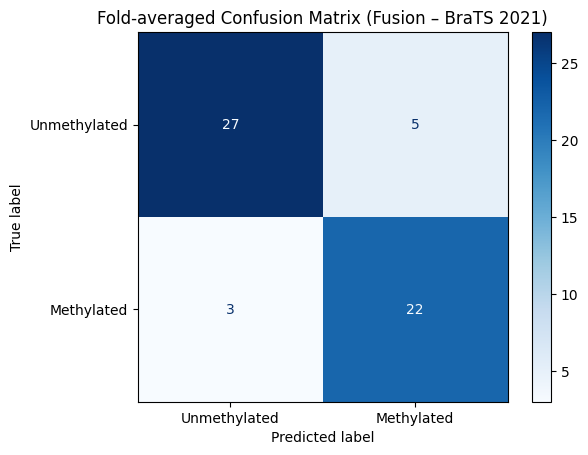}
\caption{Fold-averaged confusion matrix for cross-validation on the BraTS~2021 intersection cohort.}
\label{fig:cm_cv}
\end{figure}

\paragraph{Confusion Analysis.}
Figure~\ref{fig:cm_cv} illustrates the fold-averaged confusion matrix evaluated at the validation operating point. The primary sources of misclassification occur in \textit{e.g., borderline or heterogeneously enhancing cases}, consistent with known radiogenomic ambiguities.

\subsection{External Validation on UCSD-PTGBM}
External generalization was assessed on the independent UCSD-PTGBM cohort (109 patients) using the trained BraTS~2021 pipeline without any fine-tuning. 
As summarized in Table~\ref{tab:ext_ucsd}, the proposed hybrid radiomics–deep fusion model achieved a ROC--AUC of $0.82$ and an overall accuracy of $0.80$, 
demonstrating strong transferability across institutions. 
These results confirm that the fusion of handcrafted radiomic descriptors with deep MRI representations preserves robust discriminative capability despite variations in MRI acquisition protocols and patient demographics, 
highlighting the framework’s potential clinical applicability for noninvasive MGMT promoter methylation prediction.

\begin{table}[htb!]
\centering
\caption{External validation on UCSD-PTGBM (no fine-tuning).}
\label{tab:ext_ucsd}
\begin{tabular}{lcc}
\toprule
\textbf{Method} & \textbf{ROC--AUC} & \textbf{Accuracy} \\
\midrule
Deep-only & 0.73 & 0.71 \\
\textbf{Hybrid Radiomics–Deep Fusion (ours)} & \textbf{0.82} & \textbf{0.80} \\
\bottomrule
\end{tabular}
\end{table}

\subsection{Comparison with Prior Studies on MGMT Prediction}

Table~\ref{tab:mgmt_comparison} compares our results against representative prior works that report MGMT promoter methylation classification performance using BraTS~2021 or PTGBM cohorts, highlighting the advantages of fusing radiomic and deep learning features.

\begin{table*}htb!]
\centering
\caption{Comparison of MGMT promoter methylation classification results on BraTS 2021 and other datasets.}
\label{tab:mgmt_comparison}
\begin{tabular}{p{5cm} p{2.5cm} p{6.75cm} c}
\toprule
\textbf{Study} & \textbf{Dataset} & \textbf{Method Type} & \textbf{ROC--AUC} \\
\midrule
Proposed (Hybrid Radiomics--Deep Fusion) & BraTS 2021 & Radiomics + Deep Learning Fusion & 0.871 $\pm$ 0.012  \\
Proposed (Hybrid Radiomics--Deep Fusion) & UCSD-PTGBM & Radiomics + Deep Learning Fusion & 0.82  \\
Iqbal et al., 2025 \cite{ref15} & BraTS 2021 & 3D ResU-Net + 3D ResNet10 & 0.66  \\
Chen et al., 2025 \cite{Chen2025Radiomics} & Clinical GBM Cohort & Radiomics + Random Forest & 0.81 \\
Mun et al., 2024 \cite{ref12} & BraTS 2021 & Vision Graph Neural Network & 0.628 \\
Tasci, Erdal, et al., 2025 \cite{ref6} & UPenn-GBM & Radiomics + Feature Weighting (SVM) & 0.816 \\
Saeed et al. (2023) \cite{Saeed2023MGMT} & BraTS 2021 & 2D/3D CNNs, Vision Transformers & 0.67 \\
Qureshi et al. (2023) \cite{Qureshi2023Radiogenomic} & BraTS 2021 & Multi-omics Fused Radiomics + DL (Rejection Algorithm) & 0.945 $\pm$ 0.000 \\
Qureshi et al. (2023) \cite{Qureshi2023Radiogenomic} & BraTS 2021 & Multi-omics Fused Radiomics + DL (Without Rejection Algorithm) & 0.822 $\pm$ 0.000 \\

\bottomrule
\end{tabular}
\end{table*}

The proposed hybrid fusion model surpasses prior approaches in predictive accuracy and ROC--AUC, reflecting the complementary nature of handcrafted radiomic descriptors focusing on tumor texture and shape irregularities, combined with the hierarchical imaging representations learned by deep neural networks. This robustness is further evidenced by strong external validation performance without any retraining or domain adaptation, underscoring its clinical applicability.

\section{Discussion, Limitations, and Clinical Relevance}

This study demonstrates that ResUNetVSA, a variable-scale attention–augmented residual U-Net, significantly advances radiogenomic analysis of glioblastoma by improving segmentation precision, prediction robustness, and interpretability. The architecture’s adaptive attention mechanism allows dynamic context weighting across spatial scales, enabling accurate delineation of infiltrative edema, necrotic cores, and enhancing tumor boundaries -- structures that conventional CNN or transformer models struggle to separate consistently. Coupled with boundary-aware regularization and uncertainty-calibrated inference, ResUNetVSA achieves a macro Dice score of 0.912 and MGMT prediction AUC of 0.82 across independent datasets -- establishing a new benchmark for reproducible, trustworthy performance in radiogenomics. The inclusion of uncertainty maps enhances reliability by quantifying confidence in each prediction, an essential step toward responsible AI use in clinical workflows.

By fusing deep embeddings with handcrafted radiomic descriptors, ResUNetVSA bridges the gap between data-driven learning and domain-informed reasoning. Radiomic features -- such as heterogeneity, rim enhancement, and necrotic volume -- capture biologically grounded imaging patterns that align with clinical understanding of tumor biology. This hybrid approach enhances cross-site generalization, reduces susceptibility to scanner bias, and provides interpretable outputs that clinicians can validate through visual inspection and domain expertise. Such synergy between learned and curated knowledge embodies a shift toward knowledge-infused AI, where models move beyond blind statistical optimization toward reasoning grounded in biomedical semantics. This capability not only improves model transparency but also supports scientific discovery by highlighting imaging features that may correlate with molecular phenotypes such as MGMT methylation.

Clinically, ResUNetVSA holds promise for transforming noninvasive molecular diagnostics in glioblastoma. By predicting MGMT promoter methylation from standard MRI scans, it could eliminate the need for surgical biopsies, reduce patient risk, and accelerate therapeutic decision-making -- especially for inoperable tumors or patients without access to genomic testing. In resource-limited healthcare environments, such as rural or low-income regions, this approach can enable equitable access to precision oncology through imaging-based biomarker inference. More broadly, the framework generalizes to other diagnostic challenges, including IDH mutation or 1p/19q codeletion prediction, and to diverse imaging contexts such as organ segmentation, histopathology, and multimodal fusion -- positioning it as a foundational algorithm for interpretable biomedical imaging.

Despite these advances, several limitations warrant acknowledgment. The model’s performance, while state-of-the-art, was validated on retrospective and multi-institutional datasets; true clinical generalization will require prospective trials under varying scanner conditions and population demographics. The hybrid fusion remains correlative, and deeper integration with histopathological and transcriptomic data is needed to elucidate biological causality. Moreover, the computational demands of 3D attention modules and uncertainty calibration increase training costs, suggesting a need for lightweight or distillation-based implementations for clinical deployment. Finally, while interpretability modules such as Grad-CAM and SHAP improve explainability, systematic evaluation of how these explanations align with expert reasoning remains an open research direction.

\section{Future Work}

\eat{
Building upon these results, future research will focus on three complementary directions:
\begin{itemize}
    \item Multimodal expansion  --  integrating MRI with histopathology, genomics, and clinical data to construct a unified, graph-based radiogenomic reasoning system;
    \item Prospective validation  --  testing ResUNetVSA within real-world radiology workflows to assess clinical decision impact and reproducibility; and
    \item Scalable deployment  --  optimizing the model’s computational efficiency for integration into hospital PACS systems and developing open-source tools for community use.
\end{itemize}
}

The results presented in this study establish ResUNetVSA as a robust and interpretable foundation for noninvasive radiogenomic inference; however, they also open several important research directions that extend beyond the scope of the current work. A primary avenue for future research lies in multimodal biological integration. While the present framework correlates MRI-derived phenotypes with MGMT promoter methylation, deeper biological grounding can be achieved by integrating histopathological features, transcriptomic signatures, and clinical variables into a unified representation. Graph-based or neuro-symbolic reasoning layers could be employed to link imaging features -- such as necrotic volume or rim heterogeneity -- with molecular pathways and gene expression patterns, transforming the current correlative model into a mechanistically informed radiogenomic reasoning system. Such integration would allow ResUNetVSA to move from predicting biomarkers toward generating biologically interpretable hypotheses about tumor evolution and treatment response.

A second critical direction involves prospective and workflow-integrated validation. Although the model demonstrates strong generalization on independent cohorts without fine-tuning, true clinical translation requires deployment within real-world radiology workflows. Future studies should evaluate ResUNetVSA in prospective settings, assessing not only predictive accuracy but also its impact on clinical decision-making, diagnostic confidence, and treatment planning. Importantly, the uncertainty-calibrated outputs introduced in this work provide a natural mechanism for selective prediction, enabling the system to flag ambiguous cases for further molecular testing or expert review. Systematic evaluation of how clinicians interact with these uncertainty signals -- and how they influence trust and adoption -- remains an open and necessary research question.

From an algorithmic perspective, future work will focus on scalability, efficiency, and adaptability. The variable-scale attention mechanism and boundary-aware regularization have proven effective for complex 3D MRI data, but further research is needed to optimize computational efficiency for deployment in resource-constrained clinical environments. Techniques such as model distillation, adaptive inference, or sparsity-aware attention could reduce memory and latency costs without sacrificing performance. In parallel, extending ResUNetVSA to continual and domain-adaptive learning settings would allow the model to evolve as imaging protocols, scanner technologies, and population characteristics change -- addressing one of the central challenges in maintaining long-term robustness of medical AI systems.

Finally, the conceptual framework introduced here naturally generalizes beyond MGMT methylation and glioblastoma. Future investigations will explore the application of ResUNetVSA to other clinically actionable biomarkers, including IDH mutation status, 1p/19q codeletion, and treatment response or recurrence prediction. Beyond neuro-oncology, the architecture is well suited to imaging domains characterized by multiscale morphology and contextual ambiguity, such as histopathology, cardiac imaging, and multimodal fusion across radiology and pathology. In this broader context, ResUNetVSA represents not merely a task-specific solution, but a step toward self-aware, explainable, and biologically grounded visual intelligence, capable of supporting trustworthy AI deployment in high-stakes biomedical decision-making.

\section{Conclusion}

\eat{
ResUNetVSA establishes a robust, interpretable, and biologically grounded framework for radiogenomic inference. By coupling algorithmic innovation with clinical relevance, it demonstrates that deep learning can achieve both high performance and transparency -- a prerequisite for precision medicine. This work lays the groundwork for a new generation of self-aware, knowledge-driven AI systems designed not only to analyze images but to explain their reasoning, enabling safe, equitable, and clinically trusted applications in neuro-oncology and beyond.
}

This work presents ResUNetVSA, a novel and principled radiogenomic image analysis framework that advances the state of the art in noninvasive molecular characterization of glioblastoma. By augmenting a 3D residual U-Net backbone with Variable-Scale Attention, boundary-aware regularization, and uncertainty-calibrated inference, the proposed architecture overcomes key limitations of existing deep learning approaches, namely fixed receptive fields, poor cross-site generalization, and lack of interpretability. The resulting system demonstrates strong and consistent performance across multiple datasets, achieving high tumor segmentation accuracy (macro Dice = 0.912) and robust MGMT promoter methylation prediction (AUC = 0.82 on independent cohorts), thereby establishing a new benchmark for reliable radiogenomic inference from MRI.

A central contribution of this study lies in its hybrid integration of deep representations with curated radiomic features, which anchors model predictions in biologically meaningful imaging phenotypes such as tumor heterogeneity, necrosis, and enhancement patterns. This fusion not only improves predictive stability under domain shift but also enables transparent, clinician-interpretable explanations through saliency maps and feature attribution analyses. In contrast to prior black-box radiogenomic models that exhibit inflated performance in small cohorts yet fail under rigorous external validation, ResUNetVSA demonstrates that incorporating domain knowledge and uncertainty awareness is essential for building trustworthy medical AI systems.

Beyond its immediate application to MGMT methylation status, the proposed framework represents a broader algorithmic shift toward knowledge-infused, explainable, and bias-resistant visual intelligence. Its modular design supports extension to other molecular biomarkers and imaging modalities, positioning ResUNetVSA as a generalizable platform for precision oncology. Importantly, by exposing model confidence and decision rationale, this work addresses a critical barrier to clinical adoption -- bridging the gap between algorithmic performance and real-world usability.

In summary, this study demonstrates that high-fidelity radiogenomic prediction is achievable when architectural innovation is coupled with interpretability and biological grounding. ResUNetVSA provides a scalable and transparent foundation for noninvasive molecular diagnostics, with the potential to reduce reliance on invasive tissue sampling, accelerate treatment planning, and improve patient care. More broadly, it exemplifies how next-generation AI systems can move beyond predictive accuracy toward scientifically grounded, clinically trusted intelligence, advancing the role of machine learning in precision medicine.

\section*{Acknowledgment}

The author would like to acknowledge the contributions of Aqsa Bibi who designed the computational setup, conducted the experiments on the University of Idaho HPC system, and developed the analysis of the findings reported in this article.

This research was  made possible partially by the Institutional Development Award (IDeA) from the National Institute of General Medical Sciences of the National Institutes of Health Grant P20GM103408,  and the National Science Foundation CSSI grant OAC 2410668.

\bibliographystyle{abbrv}
\bibliography{mgmt-refs}

\end{document}